\documentclass{article}

\usepackage{arxiv}

\usepackage[utf8]{inputenc} 
\usepackage[T1]{fontenc}    
\usepackage{hyperref}       
\usepackage{url}            
\usepackage{booktabs}       
\usepackage{amsfonts}       
\usepackage{nicefrac}       
\usepackage{graphicx}
\usepackage{amsmath}
\usepackage{subcaption}
\usepackage{algorithm}
\usepackage{algpseudocode}
\usepackage[round]{natbib}

\title{Homogeneous Learning: Self-Attention Decentralized Deep Learning}
\date{} 
\author{Yuwei Sun\\
	Graduate School of Information Science and Technology\\
	University of Tokyo\\
	Tokyo, Japan 113-8654 \\
	\texttt{ywsun@g.ecc.u-tokyo.ac.jp} \\
	\And
	Hideya Ochiai \\
	Graduate School of Information Science and Technology\\
	University of Tokyo\\
	Tokyo, Japan 113-8654 \\
	\texttt{ochiai@elab.ic.i.u-tokyo.ac.jp} \\
}

\renewcommand{\headeright}{Preprint. Under review.}
\renewcommand{\undertitle}{}
\renewcommand{\shorttitle}{Homogeneous Learning}

\begin{document}
\maketitle

\begin{abstract}
Federated learning (FL) has been facilitating privacy-preserving deep learning in many walks of life such as medical image classification, network intrusion detection, and so forth. Whereas it necessitates a central parameter server for model aggregation, which brings about delayed model communication and vulnerability to adversarial attacks. A fully decentralized architecture like Swarm Learning allows peer-to-peer communication among distributed nodes, without the central server. One of the most challenging issues in decentralized deep learning is that data owned by each node are usually non-independent and identically distributed (non-IID), causing time-consuming convergence of model training. To this end, we propose a decentralized learning model called Homogeneous Learning (HL) for tackling non-IID data with a self-attention mechanism. In HL, training performs on each round’s selected node, and the trained model of a node is sent to the next selected node at the end of each round. Notably, for the selection, the self-attention mechanism leverages reinforcement learning to observe a node’s inner state and its surrounding environment’s state, and find out which node should be selected to optimize the training. We evaluate our method with various scenarios for an image classification task. The result suggests that HL can produce a better performance compared with standalone learning and greatly reduce both the total training rounds by 50.8\% and the communication cost by 74.6\% compared with random policy-based decentralized learning for training on non-IID data.  
\end{abstract}

\section{Introduction}
Decentralized deep learning (DDL) is a concept to bring together distributed data sources and computing resources while taking the full advantage of deep learning models. Nowadays, DDL such as Federated Learning (FL) \citep{fl} has been offering promising solutions to social issues surrounding data privacy, especially in large-scale multi-agent learning. These massively distributed nodes can facilitate diverse use cases, such as industrial IoT \citep{10}, environment monitoring with smart sensors \citep{11}, human behavior recognition with surveillance cameras \citep{12}, connected autonomous vehicles control \citep{13, 14}, federated network intrusion detection \citep{sun, 16}, and so forth.

Though FL has been attracting great attention due to the privacy-preserving architecture, recent years' upticks in adversarial attacks cause its hardly guaranteed trustworthiness. FL encounters various threats, such as backdoor attacks \citep{38,39,40}, information stealing attacks \citep{57}, and so on. On the contrast, fully decentralized architectures like Swarm Learning (SL) \citep{sl} leverages the blockchain, smart contract, and other state-of-the-art decentralization technologies to offer a more practical solution. Whereas, a great challenge of it has been deteriorated performance in model training with non-independent identically distributed (non-IID) data, leading to extremely increased time of model convergence. 

\textbf {Our contributions.} To this end, we propose a self-attention decentralized deep learning model called Homogeneous Learning (HL). HL leverages a shared communication policy for adaptive model sharing among nodes. A starter node initiates a training task and by iteratively sending the trained model and performing training on each round's selected node, the starter node's model is updated for achieving the training goal. Notably, a node selection decision is made by reinforcement learning agents based on the current selected node's inner state and outer state of its surrounding environment to maximize a reward for moving towards the training goal. Finally, comprehensive experiments and evaluation results suggest that HL can accelerate the model training on non-IID data with 50.8\% fewer total training rounds and reduce the associated communication cost by 74.6\%.

\textbf {Paper outline.} This paper is organized as follows. Section 2 demonstrates the most recent work about DDL and methodologies for tackling data heterogeneity problems in model training. Section 3 presents the technical underpinnings of Homogeneous Learning, including the privacy-preserving decentralized learning architecture and the self-attention mechanism using reinforcement learning agents. Section 4 demonstrates comprehensive experimental evaluations. Section 5 concludes the paper and gives out future directions of this work.

\section{Related Work}
In recent years, lots of DDL architectures have been proposed leveraging decentralization technologies such as the blockchain and ad hoc networks. For instance, \cite{li} presented a blockchain-based decentralized learning framework based on the FISCO blockchain system. They applied the architecture to train AlexNet models on the FEMNIST dataset. Similarly, \cite{lu} demonstrated a blockchain empowered secure data sharing architecture for FL in industrial IoT. Furthermore, \cite{mowla} proposed a client group prioritization technique leveraging the Dempster-Shafer theory for unmanned aerial vehicles (UAVs) in flying ad-hoc networks. 

On the other hand, methodologies for tackling data heterogeneity in DDL, especially FL, have been studied for a long time. For example, \cite{sener} presented the K-Center clustering algorithm which aims to find a representative subset of data from a very large collection such that the performance of the model based on the small subset and that based on the whole collection will be as close as possible. Moreover, \cite{wang} demonstrated reinforcement learning-based client selection in FL, which counterbalances the bias introduced by non-IID data thus speeding up the global model's convergence. \cite{sun} proposed the Segmented-FL to tackle heterogeneity in massively distributed network intrusion traffic data, where clients with highly skewed training data are dynamically divided into different groups for model aggregation respectively at each round. Furthermore, \cite{zhao} presented a data-sharing strategy in FL by creating a small data subset globally shared between all the clients. Likewise, \cite{jeong} proposed the federated augmentation where each client augments its local training data using a generative neural network.

Different from the aforementioned methodologies, our work is aimed to offer a practical solution to 1) peer-to-peer decentralized deep learning; 2) a self-attention mechanism that optimizes the node selection using reinforcement learning; 3) communication cost-aware model sharing in decentralized learning.

\section{Homogeneous Learning}
\subsection{Data Privacy and Decentralized Deep Learning}
Centralized deep learning in high performance computing (HPC) environments has been facilitating the advancement in various areas such as drug discovery, disease diagnosis, cybersecurity, and so on. Despite its broad applications in many walks of life, the associated potential data exposure of training sources and privacy regulation violation have greatly decreased the practicality of such centralized learning architecture. In particular, with the promotion of GDPR \citep{EU}, data collection for centralized model training has become more and more difficult. 

For this reason, Google proposed federated learning (FL) to alleviate the limitation of model training on distributed data. FL allows a client to train its own model based on a local dataset and achieve a better performance by sharing the training result with others, whereas without sharing the raw training data. FL quickly acquired intense attention from lots of fields related to sensitive data processing including medical image classification, face recognition, intrusion detection, finance data analysis, and so forth.

Moreover, a fully decentralized deep learning architecture is peer-to-peer networking of nodes based on decentralization and security technologies such as the token-exchange, a service capable of validating and issuing security tokens to enable nodes to obtain appropriate access credentials for exchanging resources without the central server. In this case, each node owns a local training model and performs both the function of the client and the server based on a shared communication policy, which is different from FL where the central server plays the key role in model sharing. 

\subsection{Edge Heterogeneity in Decentralized Deep Learning}
The challenges related to edge heterogeneity mainly refer to two categories, i.e., data heterogeneity (non-IID data) and hardware heterogeneity (bandwidth, device memory, computation capability, etc.). Notably, data heterogeneity usually brings about time-consuming convergence of model training, and hardware heterogeneity usually causes an increase in the waiting time of training. HL is aimed to tackle model training under the assumption of data heterogeneity and improve communication efficiency of model sharing.  

To define data heterogeneity, given a \textit{N}-node scenario where each node possesses \textit{m} non-IID data samples, $\alpha$\% of the set from a main data class and the left $(100-\alpha)$\% randomly drawn from the other classes. Then the heterogeneity level of the system is defined as $\frac{\alpha}{100}$. For example, if a node owns 60\% of its data from the main data class and 40 \% data from any other supplemental classes, then the heterogeneity level of the system is 0.6. Moreover, we assume the main data classes of these \textit{N} nodes are different from each other. In particular, when the total data classes \textit{C} is smaller than the total nodes \textit{N}, for every $\frac{N}{C}$ nodes, we assign them the same main data class. 

Furthermore, to study the communication cost of model sharing, we adopt a relative distance $d_{i,j}$ between every two nodes as the measurement, where a smaller distance represents less communication cost. To simulate distances between nodes, we employ a symmetrical matrix $D_{i\times j}$ where the bidirectional distances between two nodes are the same and the distance to a node itself $d_{i,j|i =j}$ is zero. In addition, each distance $d_{i,j|i\ne j}$ in the matrix has a random numerical value between 0 and an upper boundary $\beta$ (Equation 1).

\begin{equation}
D_{i \times j} = \begin{pmatrix}
d_{1,1} & d_{1,2} & \cdots & d_{1,j} \\
d_{2,1} & d_{2,2} & \cdots & d_{2,j} \\
\vdots  & \vdots  & \ddots & \vdots  \\
d_{i,1} & d_{i,2} & \cdots & d_{i,j} 
\end{pmatrix}  
\,\,subject\,to:\,\,d_{i,j|i = j} = 0,\,d_{i,j} = d_{j,i}, d_{i,j|i \ne j}\in (0, \beta]
\end{equation}

Where $D_{i\times j}$ is the symmetrical matrix of distances between nodes, and $d_{i,j}$ represents the relative distance from node \textit{i} to node \textit{j}.    

\subsection{Technical Fundamentals of Homogeneous Learning}

\begin{figure}[h]
\centering
\includegraphics[width=0.6\linewidth]{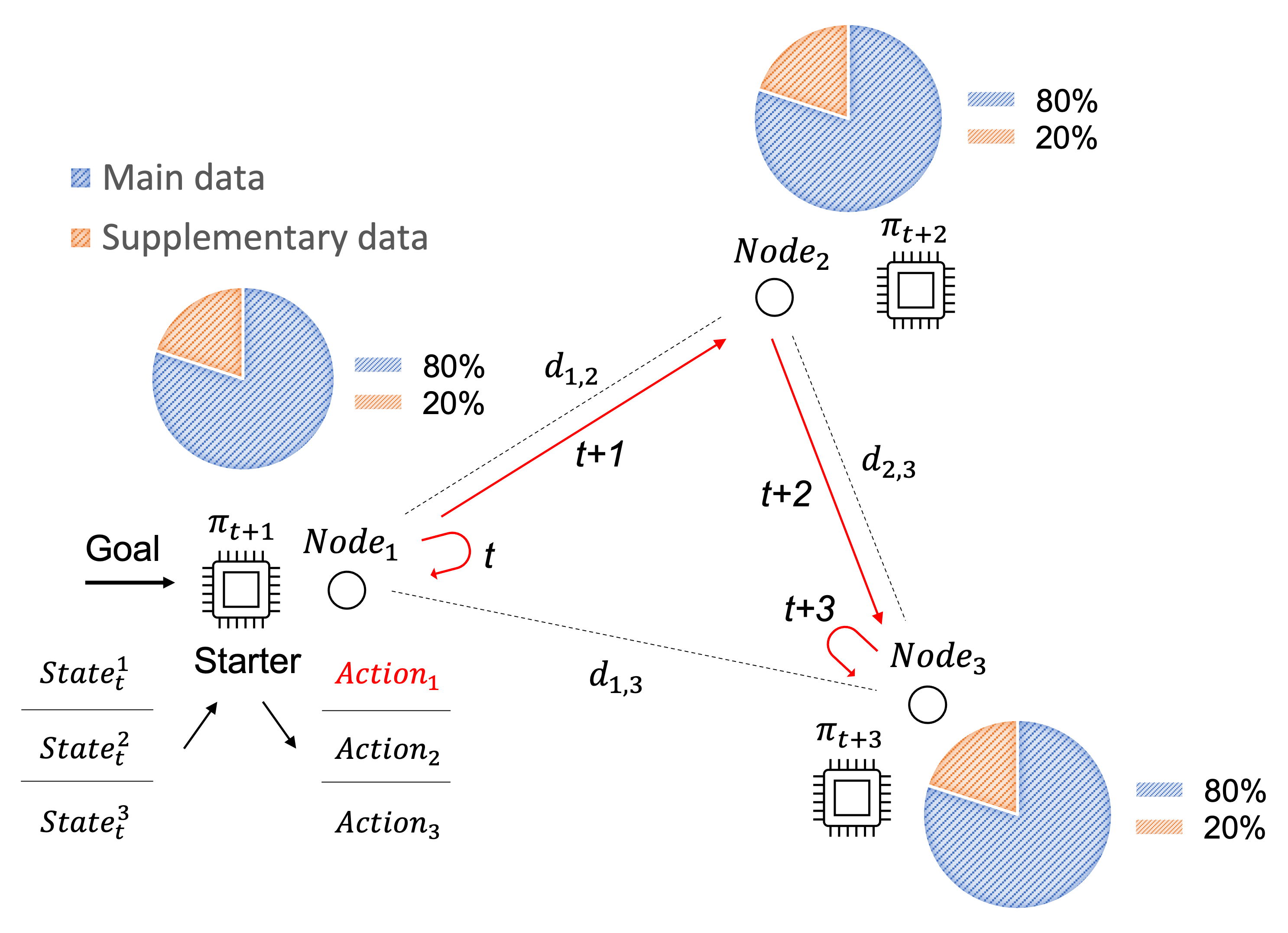}
  \caption{Homogeneous learning: self-attention decentralized deep learning.}
\end{figure}

We propose a novel decentralized deep learning architecture called Homogeneous Learning (HL) (Fig. 1). HL leverages reinforcement learning (RL) agents to learn a shared communication policy for node selection, thus contributing to fast convergence of model training and reduced communication cost for model sharing. In HL, each node has two machine learning (ML) models, namely a local foundation model for a specific ML task and an RL model for the decision-making of peer-to-peer communications. 

Moreover, we call a node that initiates a training task the starter node. First, the starter node initializes and trains a local model on its local dataset. At the same time, to measure its surrounding environment, the node requests for the model states of the other nodes in the system. After completing the training and receiving the state information, the starter node's RL agent makes the node selection decision for the next step's training based on the observation of its inner state of the trained model and the outer state from the other nodes. Then, the selected node trains the model transmitted from the former node using its local dataset. As such, the above node selection and model training will be repeated until the model performance achieves the training goal.

\subsubsection{Local Foundation Model}
In this research, we consider an image classification task, the goal of which is to successfully predict the correct labels of images owned by each node. Especially in the case of data heterogeneity, a trained model is considered to have the ability to predict images never seen before. The individual nodes in an HL system share the same model architecture which we call the local foundation model. These nodes share training results with each other to learn the underlying features that bring to a classification decision. 

As a local foundation model, we apply a three-layer convolutional neural network (CNN) model. The first convolutional layer of the CNN model has a convolution kernel of size 5×5 with a stride of 1 and it takes one input plane and it produces 20 output planes, followed by a ReLU activation function. Besides, the second convolutional layer takes 20 input planes and produces 50 output planes and it has a convolution kernel of size 5×5 with a stride of 1, followed by ReLU. Then the output is flattened followed by a linear transformation of a fully connected layer, which takes as input the tensor and outputs a tensor of size 10. Then the categorical cross-entropy is employed to compute loss. After that, we apply as a learning function the Adam to update the model based on the loss. 

\subsubsection{Reinforcement Learning Agent}
Each node in HL is associated with a reinforcement learning (RL) agent. The agent of a node aims to learn a shared communication policy by correlating the observation of the node's inner state and the outer state represented by the other nodes in the systems with the reward for an action based on the observation. As a result, the recursive self-improvement of agents through constant exploration allows faster learning of the communication policy, which we call the self-attention mechanism.

There are three main components with respect to RL, namely, the state, the action, and the reward. Based on the observation of the current state of the node and its surrounding environment, an agent selects an optimized action that will update the current state and at the same time receive a reward for the taken action. In particular, we employ a deep Q-network (DQN), which approximates a state-value function in a Q-learning framework with a neural network. Besides, we apply a DQN consisting of three fully connected layers (Fig. 2). The two hidden layers consist of 500 and 200 neurons respectively, using as an activation function the ReLU. The output layer consists of $N$ neurons representing the rewards for selecting each of the nodes in the systems. It applies a linear activation function followed by the mean squared error loss. Moreover, for each step the agent selects the node with the largest reward output. In addition, we adopt the Adam as a learning function of the DQN model. 

\begin{figure}[h]
\centering
\includegraphics[width=0.6\linewidth]{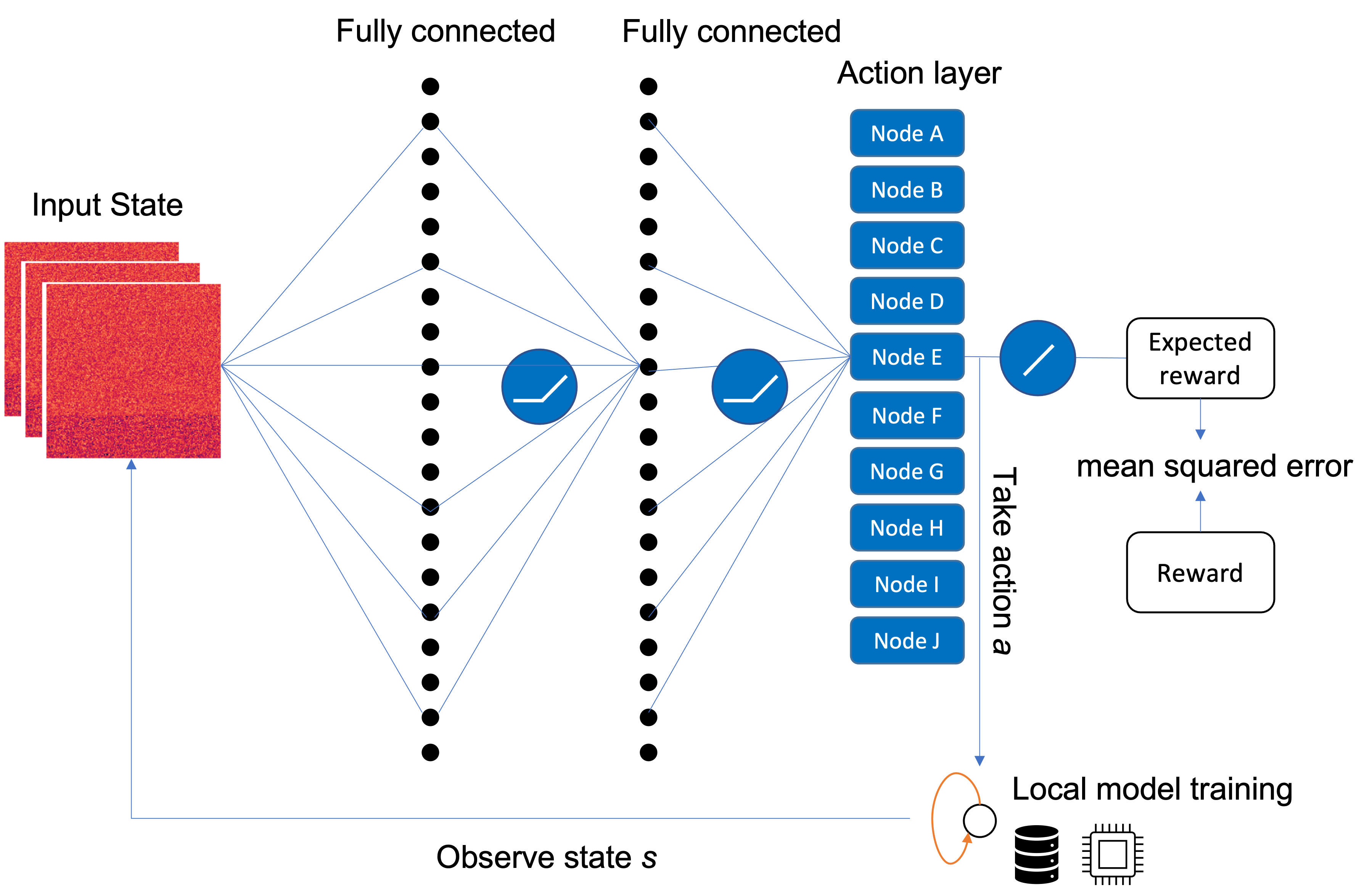}
  \caption{The self-attention mechanism based on a deep Q-network.}
\end{figure}

Moreover, to train a DQN model of a node $i$, we employ the node $i$'s model parameters $w_i$ as the inner state and the model parameters from the other nodes $w_{j|j\in N,\,j \neq i} $ as the outer state. Then, by concatenating the inner state and the outer state, it forms a feature vector representing the current state of the system. Due to the adopted foundation model of each node includes 33580 parameters, the principal component analysis (PCA) is used for dimension reduction. As a result, we convert the state feature vector from 33580 dimensions to \textit{N} dimensions, where \textit{N} is the number of total nodes in the systems. After that, based on the extracted feature vector, the DQN agent takes an action by selecting any node including itself for the next step's model training. Furthermore, every step $t$, a trained foundation model is evaluated with a hold-out validation set, and a reward $r_t$ defined in (2) is computed from the validation accuracy, the communication distance between the former node and the current node, and a penalty for taking each step. In addition, an episode refers to a whole training process of the decentralized learning, and an episode reward $R$ is defined in (3), which is the accumulative reward of the current reward and discounted future rewards.

\begin{equation}
r_{t} = 32^{ValAcc_{t}-GoalAcc}-d_{node_{t}, node_{t+1}}-1
\end{equation}
\begin{equation}
R = \sum^{T}_{t=1}\gamma^{t-1} r_t
\end{equation}

Where $ValAcc_{t}$ represents the validation accuracy at the time step \textit{t}, $GoalAcc$ is the training goal of the starter node, and $d_{node_{t}, node_{t+1}}$ is the communication distance from the symmetrical matrix $D_{i \times j}$. We employ an exponentially increasing function to distinguish between different validation results when the model is close to convergence resulting in a relatively small variance. Besides, a penalty of minus one is applied for taking each step of training.

To allow constant exploration of a DQN agent, epsilon is a factor to control the probability of a decision made by the agent at each step. In detail, for each step, a random numerical value between 0 and 1 is chosen and then compared with the latest epsilon. If the random value is greater than the epsilon, the agent will take an action based on the DQN model. In contrast, if it is less than or equal to the epsilon, the agent will take a random action instead. For either case, a sample consisting of the state, the action, the reward, and the next state will be stored in the replay memory of the agent for future policy updating. Furthermore, a progressively decaying epsilon is applied to increase the agent's decision-makings with the progress of training, defined in (4).

\begin{equation} 
Epsilon_{T+1} = Epsilon_{T}\cdot e^{- Decay}
\end{equation}

Where $Epsilon_{T+1}$ is the next episode's epsilon, $e$ is the Euler's number approximately equal to 2.71828, and $Decay$ is the epsilon decay for decreasing the epsilon every episode.

For each episode, the decentralized learning terminates when the model achieves the training goal or exceeds the maximum step of actions. Moreover, at the end of each episode, the DQN model is updated by training on a small subset of samples drawn from the replay memory, defined in (5). The updated DQN model is shared with all nodes in the systems. As a result, agents perform better and better in predicting the reward for selecting a specific node for the next round, thus improving the final episode reward and reducing the total training rounds and the communication cost.  

\begin{equation}
L(w_T^{dqn})=\sum_{i=1}^{B}l(r_i+\rho\,\underset{a}{max}\,Q(a, s^{'}_i; w_T^{dqn}), Q(a_i, s_i;  w_T^{dqn}))
\end{equation}\begin{equation*}
\theta^{*}=arg\,\underset{\theta}{min}L(\theta),\,subject\,to\,\,\theta = w_T^{dqn}
\end{equation*}

Where $w_T^{dqn}$ is the DQN model's weights, $s_i$ is the system's current state, $s^{'}_i$ is the next step's system state, $a_i$ is the taken action for the current step, $a$ is the predicted next step's action that maximizes the future reward, $\rho$ is the discount factor of the future reward, $Q$ is a feed-forward function to infer the output of the DQN model, $l$ is the mean squared error loss function, $r_i$ is the current step's reward, and \textit{B} is the training sample size from the replay memory.

The training phase of HL is defined in Algorithm 1. Algorithm 2 demonstrates the application phase of HL after obtaining the optimized communication policy for node selection.

\begin{algorithm}
\caption{Model Training of Homogeneous Learning}
\begin{algorithmic}[1]
\State Training:
\State initialize $w^{dqn}_{0}$  
\For {each episode $T = 1, 2, . . .$}
	\State initialize $w^i_0$    \Comment{$w^i_t$ is model weights of $Node_i$}
    \State $node_{t} = i$
    \For {each step $t = 0, 1, 2, . . .$}
    	\While {$ValAcc_{t} < GoalAcc$ and $t < t_{max}$} \Comment{$t_{max}$ is the maximum episode steps} 
        	  \State $ValAcc_{t}, w_{t+1}^i, w_{t+1}^{dqn} = HL(w_{t}^i, w_t^{dqn})$
              \State Send $\{w_{t+1}^i, w_{t+1}^{dqn}\} $ to $node_{t+1}$ for the next step's model update
          \EndWhile
       \EndFor
       \State $Epsilon_{T+1} = Epsilon_{T}\cdot e^{-Decay}$     
	\EndFor
\State 
\Function{HL}{$w_{t}^i, w_t^{dqn}$}
	\State $w_{t+1}^{i} = Train(w_{t}^{i}, m_i)$  \Comment{$m_i$ is $Node_i$'s local training data}
    \State $State_t^{inner} = w_{t+1}^i$  \Comment{Inner state of the node}
    \State $State_t^{outer} = \{w_t^j \mid  j \neq i, j \in N \}$ \Comment{Outer state of the surrounding environment}
    \State $s_t = Concatenate(State_t^{inner}, State_t^{outer})$
    \State $a_t = Q(s_t; w_t^{dqn}, Epsilon_{T})$
    \State $node_{t+1} = a_t $  
    \State $ r_t = 32^{ValAcc_{t}-GoalAcc}-d_{node_{t}, node_{t+1}}-1$
    \State Add $\{s_t, a_t, r_t\}$ to the replay memory
    \State $w_{t+1}^{dqn} = Train(w_t^{dqn}, B \in replay\ memory)$
    \State
    \Return $Acc_{t}, w_{t+1}^i, w_{t+1}^{dqn}$ 
\EndFunction
\end{algorithmic}
\end{algorithm}

\begin{algorithm}
\caption{Application of Homogeneous Learning}
\begin{algorithmic}[1]
\State initialize $w_0^i$
\State $node_{t} = i$
\For {each step $t = 0, 1, 2, . . .$}
	\While {$ValAcc_{t} < GoalAcc$}
        \State $w_{t+1}^{i} = Train(w_{t}^{i}, m_i)$ 
        \State $State_t^{inner} = w_{t+1}^i$
        \State $State_t^{outer} = \{w_t^j \mid  j \neq i, j \in N \}$
        \State $s_t = Concatenate(State_t^{inner}, State_t^{outer})$
        \State $a_t = Q(s_t; w^{dqn})$ 
        \State $node_{t+1} = a_t $
        \State Send $\{w_{t+1}^i, w^{dqn}\} $ to $node_{t+1}$ for the next step's model update   
     \EndWhile
\EndFor
\end{algorithmic}
\end{algorithm}

\section{Evaluation}
\subsection{Experiment Setup}
\subsubsection{Dataset} We applied MNIST \citep{mnist}, a handwritten digit image dataset containing 50,000 training samples and 10,000 test samples labeled as 0-9, to evaluate the performance of the proposed method. Then, a 10-node scenario training on the MNIST dataset was studied for achieving a validation accuracy goal of 0.80. Moreover, we conducted all experiments on a GPU server with 60 AMD Ryzen Threadripper CPUs, two NVidia Titan RTX GPUs with 24 GB RAM each, and Ubuntu 18.04.5 LTS OS. The machine learning library we used to build the system is Tensorflow. 

\subsubsection{Baseline Models} To compare the performance of the proposed approach, we considered three different model training methods as baselines, which are centralized learning with all data collected from the 10 nodes, decentralized learning with a random node selection policy, and standalone learning of the starter node on its local data without model sharing. In detail, for each method, we applied the same local foundation model architecture and associated model training hyperparameters using the training set data from MNIST. The goal is to achieve a validation accuracy of 0.80 for a trained model on the hold-out test set of the MNIST dataset.  
Besides, regarding the standalone method, we utilized the early stopping to monitor the validation loss of the model at each epoch with a patience of five, which automatically terminated the training process when there appeared no further decrease in the validation loss of the model for the last five epochs. Furthermore, in the centralized and standalone learning, evaluation was performed at each epoch of the training. On the other hand, in decentralized learning, due to multiple models in a system, the evaluation was performed on the trained local model of each step's selected node.

\subsubsection{Homogeneous Learning Settings} Homogeneous Learning (HL) of \textit{N}=10 nodes with a heterogeneity level of 0.8 was adopted. Each node \textit{i} owned a total of \textit{m}=500 non-IID data samples, 80\% of the set from a main data class $c^i_{main}$ and the left 20\% randomly drawn from the other classes $c^i_{sup}$. Moreover, the main data classes of these 10 nodes were different from each other (subject to $c_{main|i \in N}^i = i, i\in\{0,1,2,3,4,...,9\}$). Likewise, the training goal of HL was 0.80 based on the validation accuracy of a selected node's model at each round with the test set of MNIST. In addition, to generate the distance matrix, the relative communication cost represented by the distance between two different nodes $d_{i,j|i\ne j}$ takes a random numerical value between 0 and 0.1. Besides, a random seed of 0 was adopted for the reproducibility of the distance matrix (See \ref{heatmap}). 

For the local foundation model training, we adopted an epoch of one with a batch size of 32. A further discussion on the selection of these two hyperparameters can be found in the section \ref{optimization}. Moreover, the Adam was applied as an optimization function with a learning rate of 0.001.  

\subsection{Numerical Results}
\subsubsection{Communication Policy Updating Based on Deep Q-Networks}
As aforementioned, each node has a specific main data class. We considered a starter node with the main data class of digit '0'. Then, starting from the starter node, a local foundation model was trained on the current node's local data and sent to the next step's node decided by either a deep Q-network (DQN) model or a random action based on the epsilon. We adopted an initial epsilon of one and a decay rate of 0.02. Moreover, the DQN model was updated at the end of each episode using the hyperparameters defined in Table 1. We applied a maximum replay memory size of 50,000 and a minimum size of 128, where the training of the DQN model started only when there were more than 128 samples in the replay memory and the oldest samples would be removed when samples were more than the maximum capacity. Furthermore, in every episode, an agent randomly drew 32 samples from the memory to update its model, with a total of 120 episodes. A relatively small number of episodes is due to time-consuming model training associated with the policy exploration in decentralized learning.

\begin{table}[t]
\caption{Hyperparameters in Homogeneous Learning}
\begin{center}
\begin{tabular}{llll}
\multicolumn{2}{c}{\bf LOCAL FOUNDATION MODEL}  &\multicolumn{2}{c}{\bf DQN MODEL}
\\ \hline \\
Epoch    &   1       &  Episode & 120 \\
Batch size & 32            &Future reward discount & 0.9\\
Learning  rate & 0.001     &Epsilon decay & 0.02\\
Optimization function  & Adam        &Epoch & 1\\
Maximum training rounds  &  35     &Batch size & 16\\
    &       &Learning rate & 0.001\\
\end{tabular}
\end{center}
\end{table}

For each episode, we computed the step rewards and the episode reward for the model training to achieve the performance goal. With the advancement of episodes, the communication policy evolved to improve the episode reward thus benefiting better decision-making of the next-node selection. Figure 3 illustrates the episode reward and the mean reward over the last 10 episodes during the DQN model training.

\begin{figure}[h]
\centering
\includegraphics[width=0.6\linewidth]{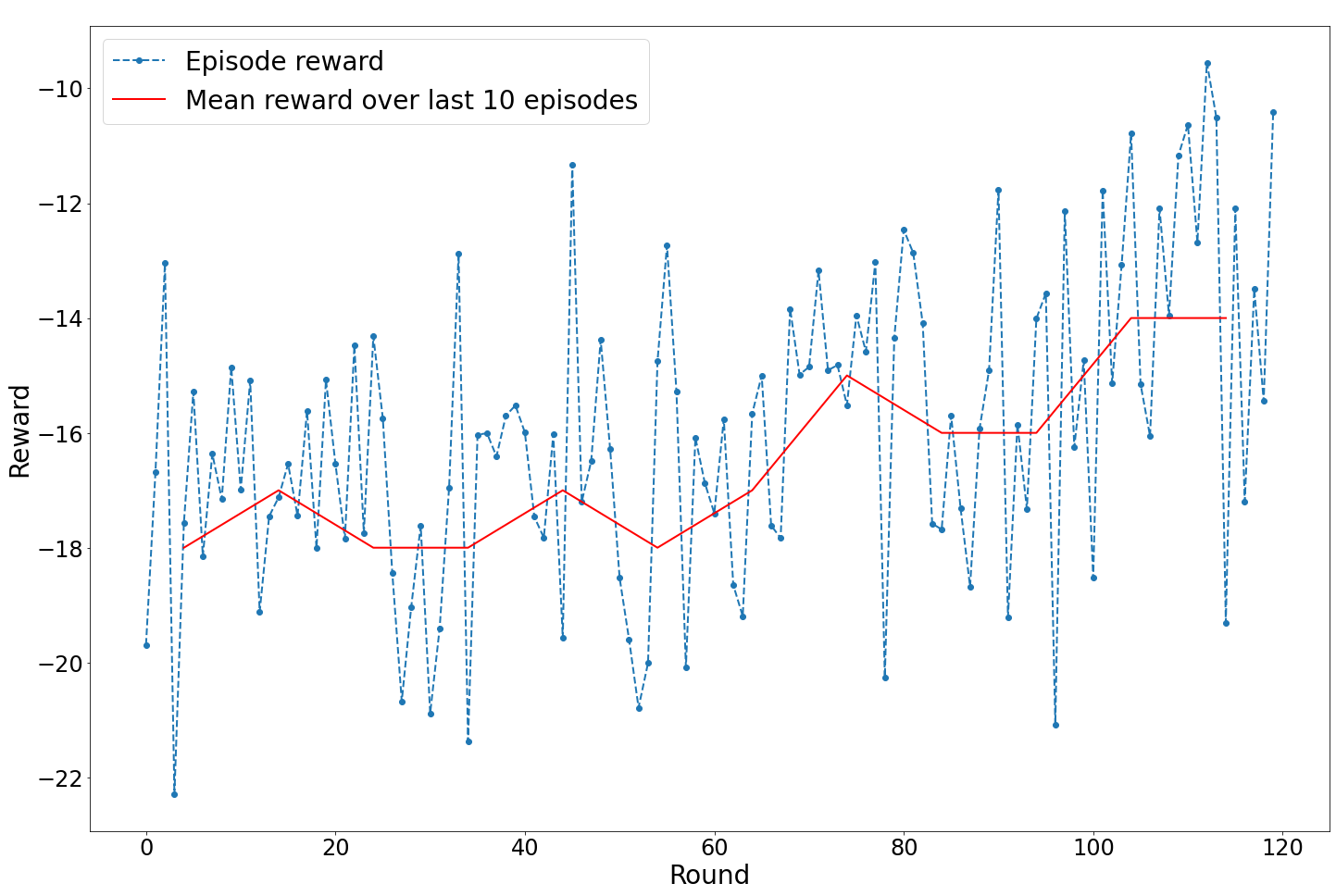}
\caption{With the increase of episodes, the mean reward over last 10 episodes is gradually increasing. The DQN model learned a better communication policy by training on samples from the replay memory, which could contribute to the systems' performance concerning required total training rounds and communication cost.}
\end{figure}

\subsubsection{Comparisons Regarding Computational and Communication Cost}
To compare the performance between HL and the aforementioned three baseline models, we performed a comprehensive evaluation against the metrics of computational cost and communication cost. Notably, the computational cost refers to the required total rounds for a system to achieve the training goal, and the communication cost refers to the total communication distance for the model sharing in decentralized learning.   

\begin{figure}[!htb]
   \begin{minipage}{0.49\textwidth}
     \centering
     \includegraphics[width=\linewidth]{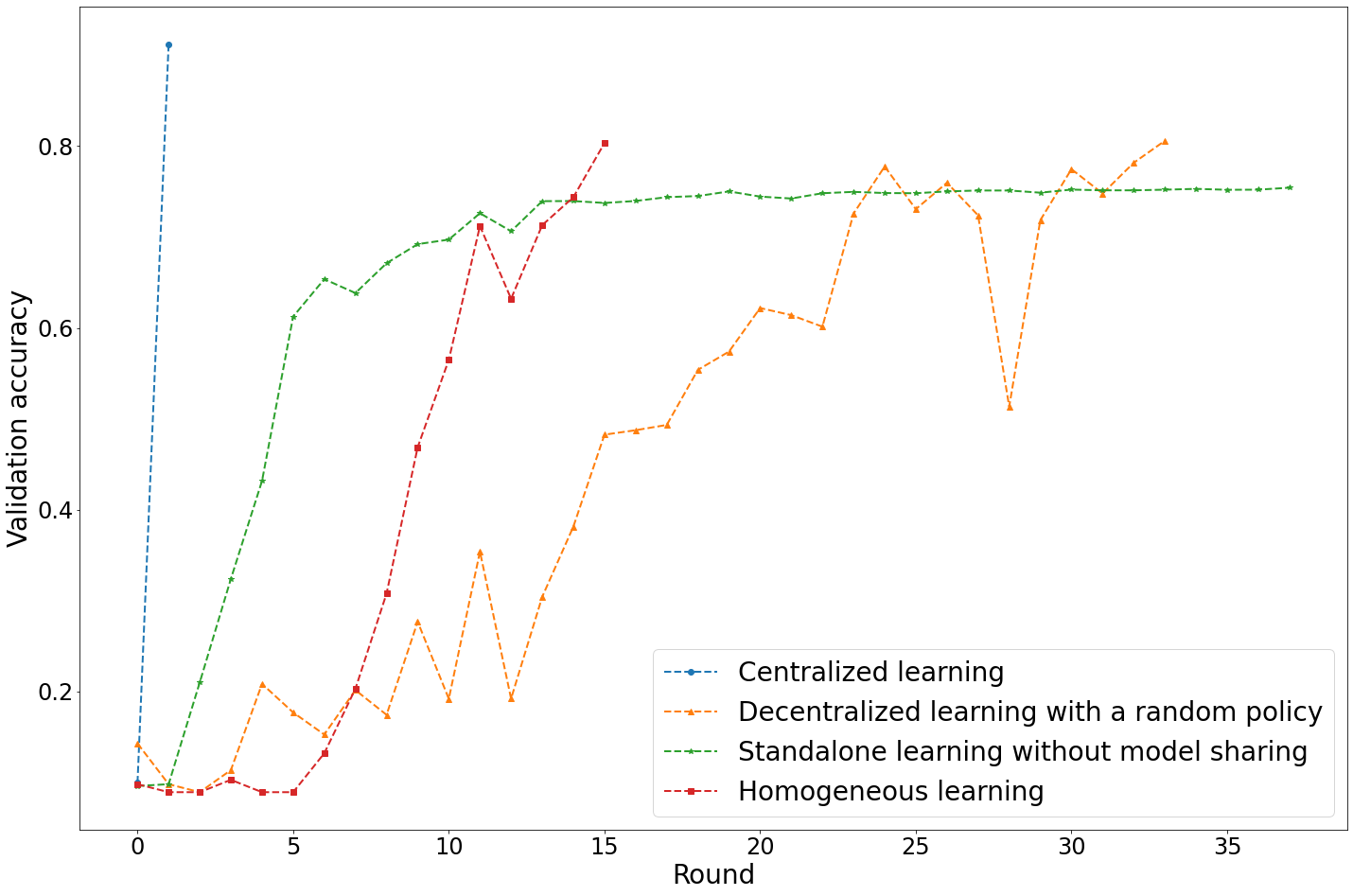}
     \caption{Validation accuracy at each round of model training based on four different methods.}
   \end{minipage}\hfill
   \begin{minipage}{0.49\textwidth}
     \centering
     \includegraphics[width=\linewidth]{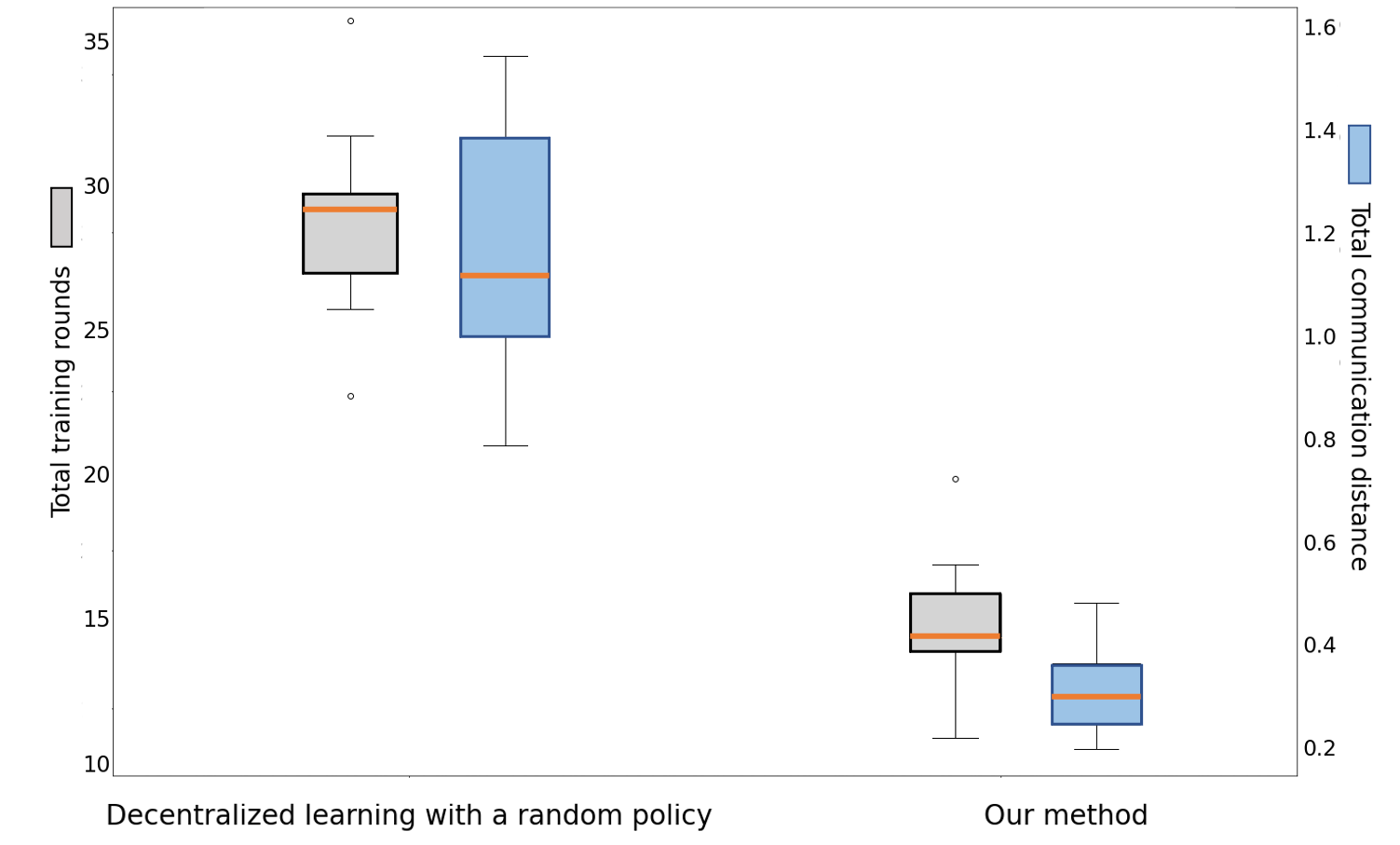}
     \caption{Performance comparison between the random policy-based decentralized learning and our method (HL). Each error bar illustrates ten individual experiments' results.}
   \end{minipage}
\end{figure}

\paragraph{Computational Cost} Figure 4 shows the total training rounds required for each method to achieve the final goal. First, for the standalone learning, due to its limited local training data, the training appeared to be extremely slow after the validation accuracy reached 0.70. Finally, it terminated with a final accuracy of around 0.75 due to the early-stopping strategy. Moreover, by comparing the decentralized learning methods with and without the self-attention mechanism, the result suggests that our proposed method of HL can greatly reduce the total training rounds. In addition, though centralized learning shows the fastest convergence, it suffers from problems of data privacy. 

\paragraph{Communication Cost} In decentralized learning, to train a model, each selected node trains the current model on its local dataset and sends the trained model to another node, and the communication cost refers to the network traffic payload for sending model weights. We studied the relative cost by introducing the communication distance between every two nodes. Then, the communication cost of a decentralized method is defined as the total communication distance for the model sharing, from the starter node to the last selected node. 

We performed ten individual experiments for each method and used as final results the best cases of node selection over the last five episodes when decisions were almost made by the agent and a learned communication policy was prone to be stable. Figure 5 illustrates the experiment results of the total training rounds and the communication cost when applying the two methods respectively. The bottom and top of the error bars represent the $25_{th}$ and $75_{th}$ percentiles respectively, the line inside the box shows the median value, and outliers are shown as open circles. Finally, the evaluation result shows that HL can greatly reduce the training rounds by 50.8\% and the communication cost by 74.6\%.

\section{Conclusion}
Decentralized deep learning (DDL) leveraging distributed data sources contributes to a better neural network model while safeguarding data privacy. Despite the broad applications of DDL models such as federated learning and swarming learning, the challenges regarding edge heterogeneity covering the data heterogeneity and device heterogeneity have greatly limited their scalability. In this research, we proposed a self-attention decentralized deep learning method of Homogeneous Learning (HL) that recursively updates a shared communication policy by observing the system's state and the gained reward for taking an action based on the observation. We comprehensively evaluated the method by comparing it with three baseline models, applying as criteria the computational and communication cost for achieving the training goal. The evaluation result shows that HL can greatly reduce the above two types of cost. In future, a decentralized model leveraging various communication policies at the same time to achieve diverse goals is considered for the further study of this research.

\bibliographystyle{unsrtnat}
\bibliography{references}

\appendix
\section{Appendix}
\subsection{Communication Distance Matrix}
\label{heatmap}
Figure 6 illustrates the generated distance matrix $D_{i \times j}$ when applying a $\beta$ value of 0.1 and a random seed of 0.

\begin{figure}[h]
\centering
\includegraphics[width=0.6\linewidth]{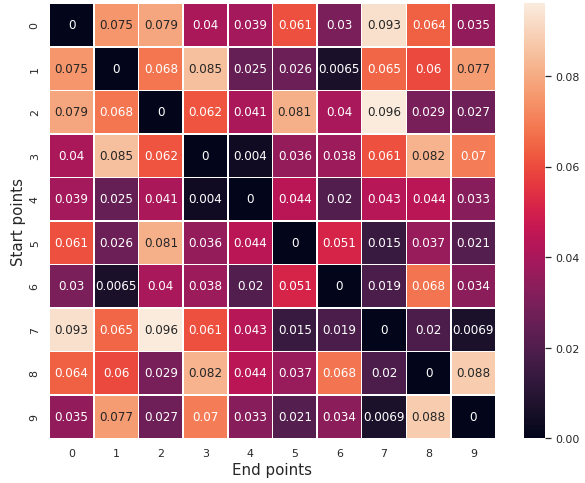}
  \caption{The adopted distance matrix $D_{i \times j}$.}
\end{figure}

\subsection{Optimization of Model Distribution Representation}
\label{optimization}
Under the assumption of data heterogeneity, to allow a reinforcement learning (RL) agent to efficiently learn a communication policy by observing model states in the systems, a trade-off between the batch size and the epoch of local foundation model training was discussed. Figure 7 illustrates the trained models' weights distribution after applying the principal component analysis (PCA), with different batch sizes and epochs applied to train on the MNIST dataset. Moreover, it shows a 100-node scenario where each color represents nodes with the same main data class. As shown in the graphs, various combinations of these two parameters have different distribution representation capabilities. By comparing the distribution density and scale, we found that when adopting a batch size of 32 and an epoch of one the models distribution was best represented, which could facilitate the policy learning of an agent. 

\begin{figure}[h]
\centering
\includegraphics[width=\linewidth]{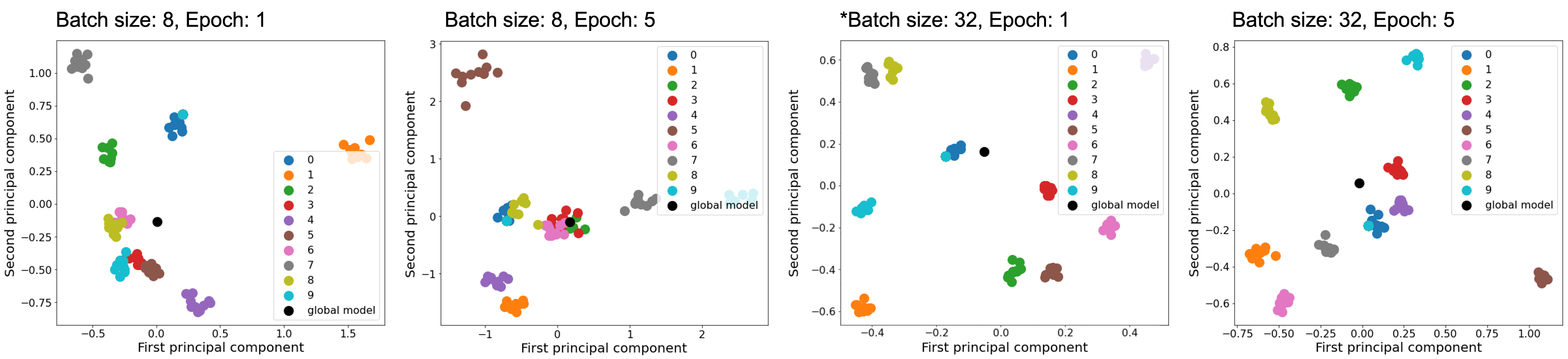}
  \caption{Optimization of model distribution representation.}
\end{figure}

\end{document}